\documentclass{llncs}
\usepackage{graphicx}
\usepackage{color}
\usepackage{subcaption}
\usepackage{wrapfig}

\begin{document}
\frontmatter          
\pagestyle{headings}  
\addtocmark{Hamiltonian Mechanics} 

\title{Serious Games Application for Memory Training Using Egocentric Images}

\author{Gabriel Oliveira-Barra\inst{1} \and Marc Bola\~{n}os\inst{1} \and Estefania Talavera\inst{1,2} \and  Adri\'{a}n Due\~{n}as\inst{1} 	\and Olga Gelonch\inst{3} \and Maite Garolera\inst{3}}
\institute{Universitat de Barcelona \and University of Groningen \and Consorci Sanitari de Terrassa }

\maketitle

 
\begin{abstract}
Mild cognitive impairment is the early stage of several neurodegenerative diseases, such as Alzheimer's. In this work, we address the use of lifelogging as a tool to obtain pictures from a patient's daily life from an egocentric point of view. We propose to use them in combination with serious games as a way to provide a non-pharmacological treatment to improve their quality of life. To do so, we introduce a novel computer vision technique that classifies rich and non rich egocentric images and uses them in serious games. We present results over a dataset composed by 10,997 images, recorded by 7 different users, achieving 79\% of F1-score. Our model presents the first method used for automatic egocentric images selection applicable to serious games.

\end{abstract}


\section{Introduction}

\begin{wrapfigure}{r}{0.4\textwidth} 
\vspace{-1.5em}
\centering
\includegraphics[width=0.4\textwidth,height=45mm]{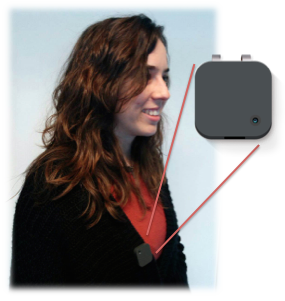} 
\vspace{-0.5em}
\caption{Person using the Narrative Clip camera.}
\label{fig:methodproposed}
\vspace{-1.5em}
\end{wrapfigure}

Dementia can result from different causes, the most common being Alzheimer’s disease (AD) \cite{fratiglioni1991prevalence}, and it is often preceded by a pre-dementia stage, known as Mild Cognitive Impairment (MCI), characterized by a cognitive decline greater than expected by an individual's age, but which does not interfere notably with their daily life activities \cite{petersen1999mild,gauthier2006mild}. Currently, medical specialists design and apply special activities that could serve as a treatment tool for cognitive capabilities enhancement. Even though, these activities are not specially designed for the patients, which limits their engagement in some cases.
 
\begin{figure}[t]
\centering
\includegraphics[width=1\textwidth,height=16mm]{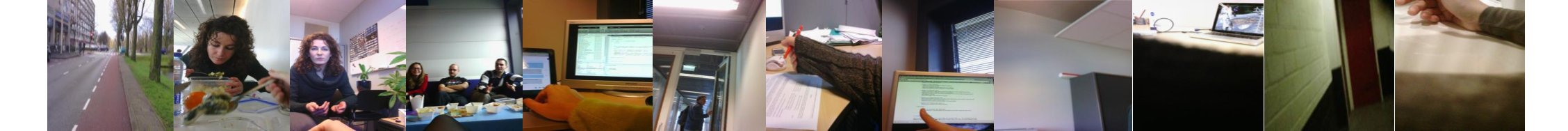}
\caption{Examples of egocentric images recorded by the Narrative Clip camera.}
\label{fig:exampleImages}
\end{figure}

A possible alternative to the application of generic exercises would be the use of personalized images of the daily life of the patients acquired by lifelogging devices. Lifelogging consists of a user continuously recording their everyday experiences, typically via wearable sensors including accelerometers and cameras, among others. When the visual signal is the only one recorded, typically by a wearable camera, it is referred to as visual lifelogging \cite{bolanos2017toward}. This is a trend that is rapidly increasing thanks to advances in wearable technologies over recent years. Nowadays, wearable cameras are very small devices that can be worn all-day long and automatically record the everyday activities of the wearer in a passive fashion, from a first-person point of view. As an example, Fig. \ref{fig:exampleImages} shows pictures taken by a person wearing such a camera.

Recent studies have described wearable cameras or lifelogging technologies as useful devices for memory support for people with episodic memory impairment, such as the one present in MCI \cite{lee2008lifelogging,doherty2012experiences}. The design of new technologies to be applied on this field requires to take into account people capabilities, limitations, needs and the acceptance of the wearable devices, since it can directly affect the treatment. So far, some studies have deeply focus into the factors associated to the use of these devices \cite{sellen2010beyond,gurrin2014lifelogging}. 

\subsubsection{Lifelogging and privacy:}
In terms of privacy, in 2011, the European Union agency ENISA evaluated the risks, threats and vulnerabilities of lifelogging applications with respect to central topics as privacy and trust issues. In their final report, they highlighted that lifelogging itself is still in its infancy but nevertheless will play an important role in the near future \cite{risk}. Therefore, they recommended further and extensive research in order to influence its evolution to “be better prepared to mitigate the risks and maximize the benefits of these technologies”. In addition, other researchers have also evaluated the possible ethical risks involved on using lifelogging devices on medical studies \cite{doherty2013wearable}.

 
\subsubsection{Serious games for MCI:}
\label{section:SeriousgamesforMCI}
Serious games (also known as games with a purpose) are digital applications specialized for purposes other than simply entertaining, such as informing, educating or enhancing physical and cognitive functions. Nowadays they are widely recognized as promising non-pharmacological tools to help assess and evaluate functional impairments of patients, as well as to aid with their treatment, stimulation, and rehabilitation \cite{robert2014recommendations}. Boosted by the publication of a Nature letter showing that video game training can enhance cognitive control in older adults \cite{anguera2013video}, there is now a growing interest in developing serious games specifically adapted to people with AD and related disorders. Preliminary evidence shows that serious games can successfully be employed to train physical and cognitive abilities in people with AD, MCI, and related disorders \cite{SGmanera2015kitchen}. \cite{mccallum2013dementia} performed a literature review of the experimental studies conducted to date on the use of serious games in neurodegenerative disorders and \cite{robert2014recommendations} studied recommendations for the use of serious games in people with AD and related disorders, reporting positive effects on several health-related capabilities of MCI patients such as voluntary motor control, cognitive functions like attention and memory or social and emotional functions. For instance they can improve their mood and increase their sociability, as well as reduce their depression.


\subsubsection{Our contribution:} 

Different studies have proven the benefits of directly stimulating the working memory. Our contribution in this paper consists in using as stimuli the autobiographical images of the MCI patients that was acquired by the wearable cameras. By doing this, we intend to accomplish the goal of enhancing their motivation and at the same time treat them in a more functional and multimodal manner \cite{flak2014memory,li2011cognitive,alves2014cognitive}.
The application, which will allow the user to exercise either at the sanitary center or at home, will be composed by serious games where the patient has to observe a series of images and interact with them.

Although the stimuli provided by egocentric images can be of greater importance than non-personal images, it is important to note both, that egocentric images are captured in an uncontrolled environment, and that wearable cameras usually have free motion that might cause most images to be blurry, dark or empty of semantic content. Considering this important limitations together with the limited capabilities of MCI patients, we propose the development of an egocentric rich images detection system intended to select only images with semantic and relevant content. 
Our hypothesis is that, by using personal daily life rich images, the motivation of the patient will increase, and as a consequence, the health-related benefits provided by the treatment. 


This paper is organized as follows. We describe the proposed serious game and model for rich images selection in Section \ref{section:seriousgames} and Section \ref{section:approachrichimages}, respectively. In Section \ref{section:results}, we describes the experimental setup and show quantitative and qualitative evaluation. Finally, Section \ref{section:conclusionsandfuture} draws conclusions and outlines future works.

\section{Proposed Serious Game: "Position Recall"}
\label{section:seriousgames}

MCI patients experiment problems in their working memory \cite{saunders2010attention}, herefore, it is of high importance to do exercises for stimulating it. All this under the neuroplasticity paradigm, which has proven that it is possible to modify the brain capabilities and the hypothesis of "use it or lose it", which are the basis of the studies related to the cognitive stimulation of elderly people \cite{salthouse2006mental}.
Thus, in this work, we introduce a serious game that we name as "Position Recall", which was designed by neuropsychologist of Consorci Sanitari de Terrassa for improving the working memory. 
The mechanics of this game follow this scheme: 

The first screen explains to the patient the instructions of the game and in the second the patient is informed that, before starting the game, there will be some practice examples that will serve to understand its logic. To start, the patient must select his preferred level of difficulty (Level 1, 2 or 3).

\begin{itemize}
\item \textbf{Level 1} shows 3 images of the patients' day during 8 seconds and they are asked to remember their positions. Immediately after they disappear, a single "target" image is shown and they are asked to select in what position it was placed. After some trials the number of images displayed are increased to 4 and then to 5.

\item \textbf{Level 2} follows the same procedure as the 1st level, but the timespan between the moment where the images disappear and the target image is shown is increased. During this timespan, called latency time, a black screen is shown.

\item \textbf{Level 3} follows the same procedure as the 2nd level, but now a distractor image is shown instead of a black screen during the latency time. The distractor image is also an image belonging to the patients' day.  
\end{itemize}

\begin{figure}[h]
 \centering
 \begin{subfigure}{.45\textwidth}
 \includegraphics[width=1\textwidth]{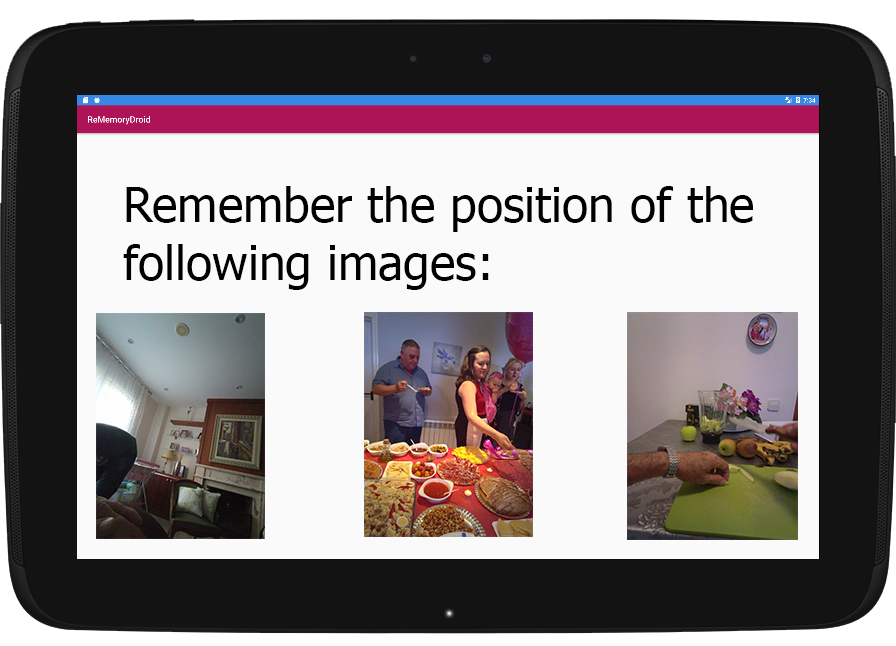}
  \caption{A predefined number of pictures of the patient is shown to him during few seconds at random positions in the screen.}
 \label{fig:aaa}
 \end{subfigure}\hfill%
 \begin{subfigure}{.45\textwidth}
 \includegraphics[width=1\textwidth]{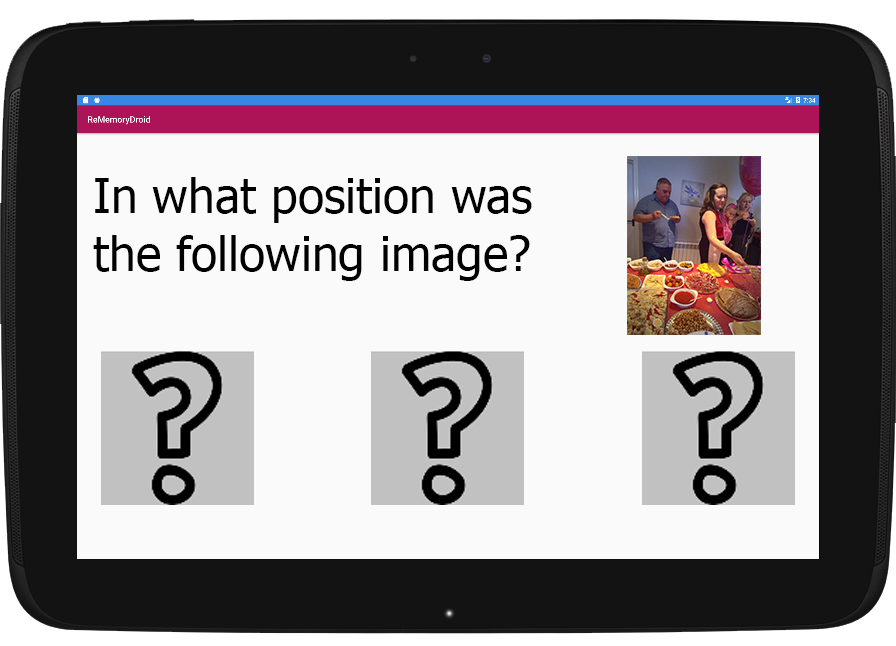} 
 \caption{After a certain time passed, the patient is asked to recall in what position one of the pictures, picked up randomly, was placed before.}
 \label{fig:bbb}
 \end{subfigure}
\end{figure}

The reward system of the game are points 
that are given after each level, and are calculated as $100 x \textit{number of correct answers}$.
There are 10 trials per level translating into a maximum of 1000 points per level and maximum of 3000 points per game. Figures \ref{fig:aaa} and \ref{fig:bbb} show the mechanics of the developed game.

The images to be shown during the serious games should be significant for the patient. We propose to use images that represent past moments of the user's life, i.e. from the egocentric photostreams recorded by the patient. On the following section, we describe the proposed model for rich images selection.


 
\section{What did I see? Rich images detection}
\label{section:approachrichimages}

The main factor for providing a meaningful image selection algorithm is the fact that the proposed serious games intend to work on cognitive and sentiment enhancement. Considering the free-motion and non-intentionality of the pictures taken by wearable cameras \cite{bolanos2017toward}, it is very important to provide a robust method for images selection.

\begin{figure}[ht!]
\centering    
\includegraphics[width=1\textwidth]{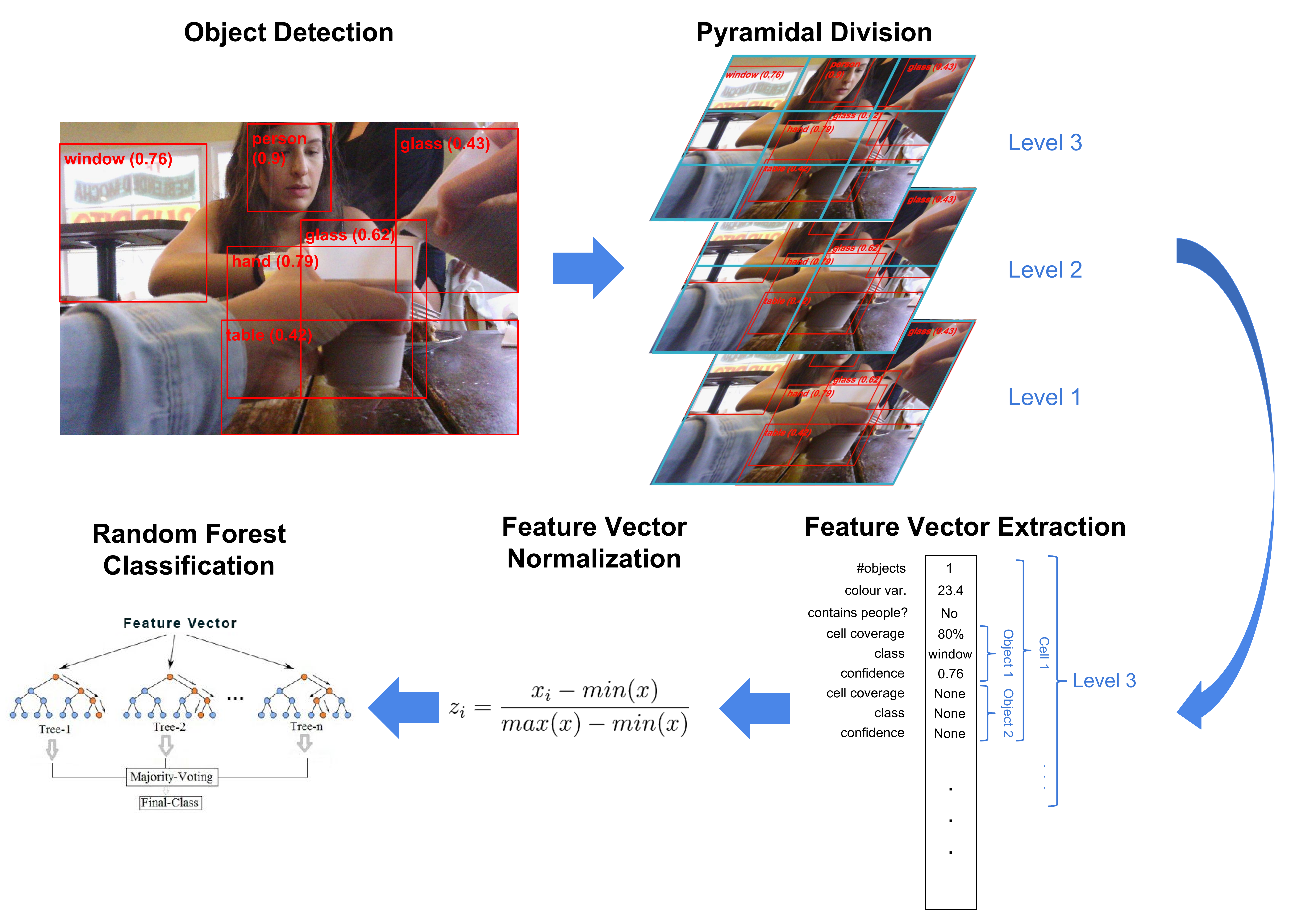}
    \caption{Scheme of the proposed rich images detection model.}
\label{fig:pipeline}
\end{figure}

Two of the most important and basic factors that determine the memorability of an image \cite{khosla2015understanding,carneegomemnet} can be described as 1) the appearance of human faces, and 2) the appearance of characteristic and recognizable objects. In this paper, we focus on satisfying the second criterion by proposing an algorithm based on computer vision. Our proposal consists in a rich images detection algorithm, which intends to detect images with a high number of objects and variability and at the same time avoids images with low semantical content, understanding as rich any image that is neither blur, nor dark and that contains clearly visible non-occluded objects. In Fig. \ref{fig:pipeline} we show the general pipeline of our proposal.

Our algorithm for rich images detection (consists in 1) objects detection: where the neural network named YOLO9000 \cite{redmon2016yolo9000} is applied in order to detect any existent object in the images and their associated confidences $c_i$. 2) the image is divided in a pyramidal structure of cells, 3) a set of richness-related features are extracted, 4) the extracted features are normalized and 5) a Random Forest Classifier (RFC) \cite{criminisi2012decision} is trained to distinguish the differences between rich or non-rich images. When extracting features, the image is divided in a pyramidal structure of cells with different sizes at each level. The set of extracted features are:

\begin{itemize}
\item \textbf{Numbers of objects the cell contains.}
\item \textbf{Variance of color in the cell.}
\item \textbf{Does the cell contain people?}
\item \textbf{Object Scale.} Real number between 0 and 1.
\item \textbf{Object Class.} Class identifier that varies between 1 and 9418.
\item \textbf{Object Confidence $c_i$.}
\end{itemize}
where all features are repeated for each cell and the last three kinds of features are repeated for each object appearing in the cells.
The image cell divisions applied are 1$x$1, 2$x$2 and 3$x$3, the maximum of objects selected per cell are 5, 3 and 2, respectively and all objects are sorted by their confidence $c_i$ before selection. If the number of objects is less than the maximum number are found, the feature value in that specific position is set to 0.

The pyramidal division of the images helps us consider smaller objects at higher levels (more cells) and bigger objects at lower levels (less cells). Thus, both small and big objects will be considered for the final prediction.

In order to define the feature "Does the cell contain people?" we manually selected a set of person-related objects detected by the employed object detection method.
The concepts representing people that we selected are "person", "worker", "workman", "employee", "consumer", "groom" and "bride".


\section{Results}
\label{section:results}

This section describes the results obtained in a quantitative and qualitative form. We compare the results obtained by variations of the proposed method on a self-made dataset of rich images.

\subsubsection{Dataset:}

The dataset used for evaluating our model was acquired by the wearable camera Narrative Clip 2\footnote{\url{www.getnarrative.com}}, which takes a picture every thirty seconds automatically. The camera was worn during 15 days by 7 different people. Considering that on average the camera takes 1,500 images per day, our dataset consists of 10,997 photographs. 

The resulting data was labeled by neuropsychologist experts on MCI cognition following the criteria that any rich image has to be 1) properly illuminated, 2) not blurry and 3) contain one or more objects that are not occluded. After this manual selection the acquired images where split in 6,399 rich images and 4,598 non-rich images.

\begin{figure}[h]
 \centering
 \begin{subfigure}{.5\textwidth}
 \includegraphics[width=.38\textwidth]{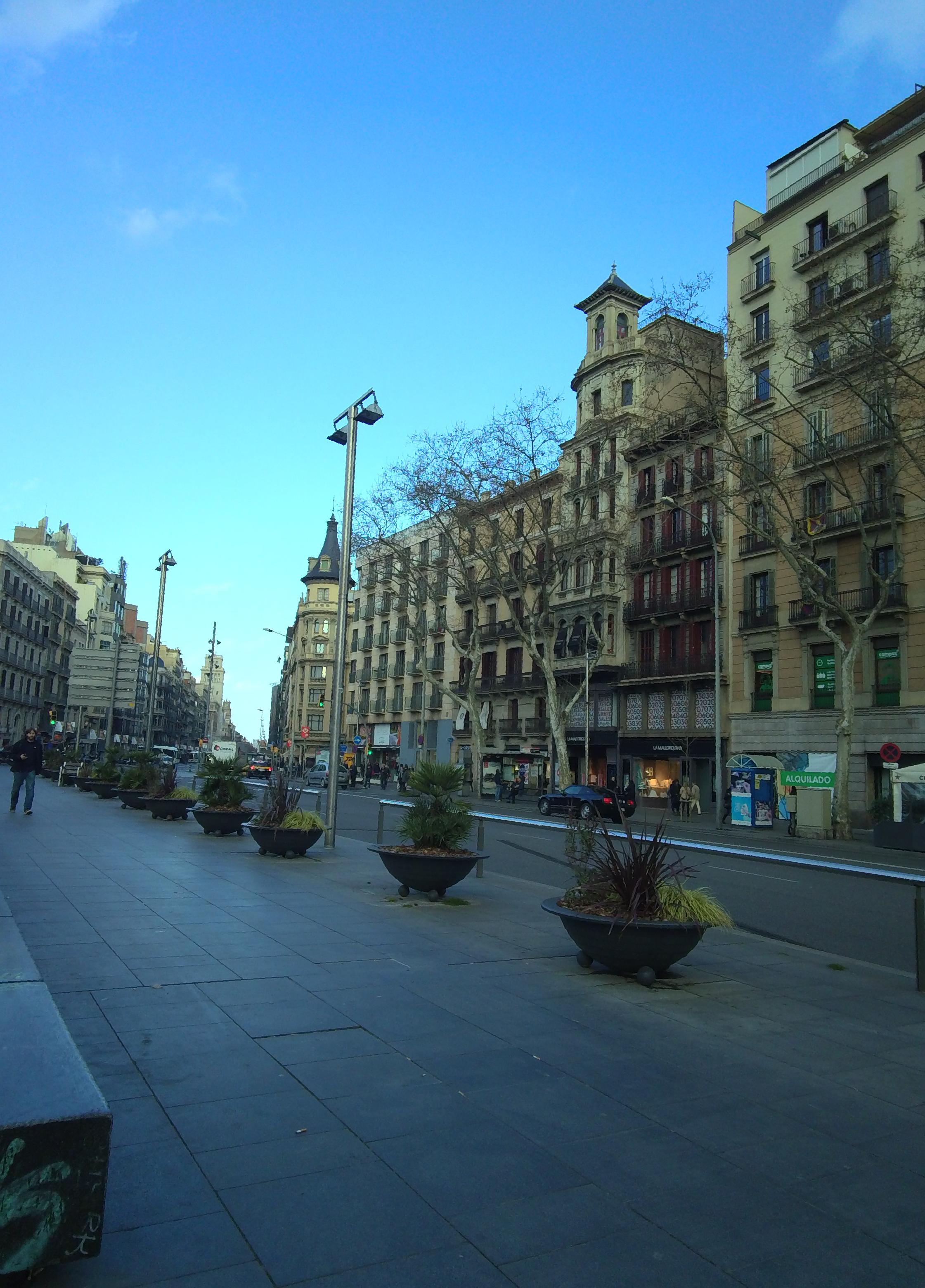}
 \includegraphics[width=.49\textwidth]{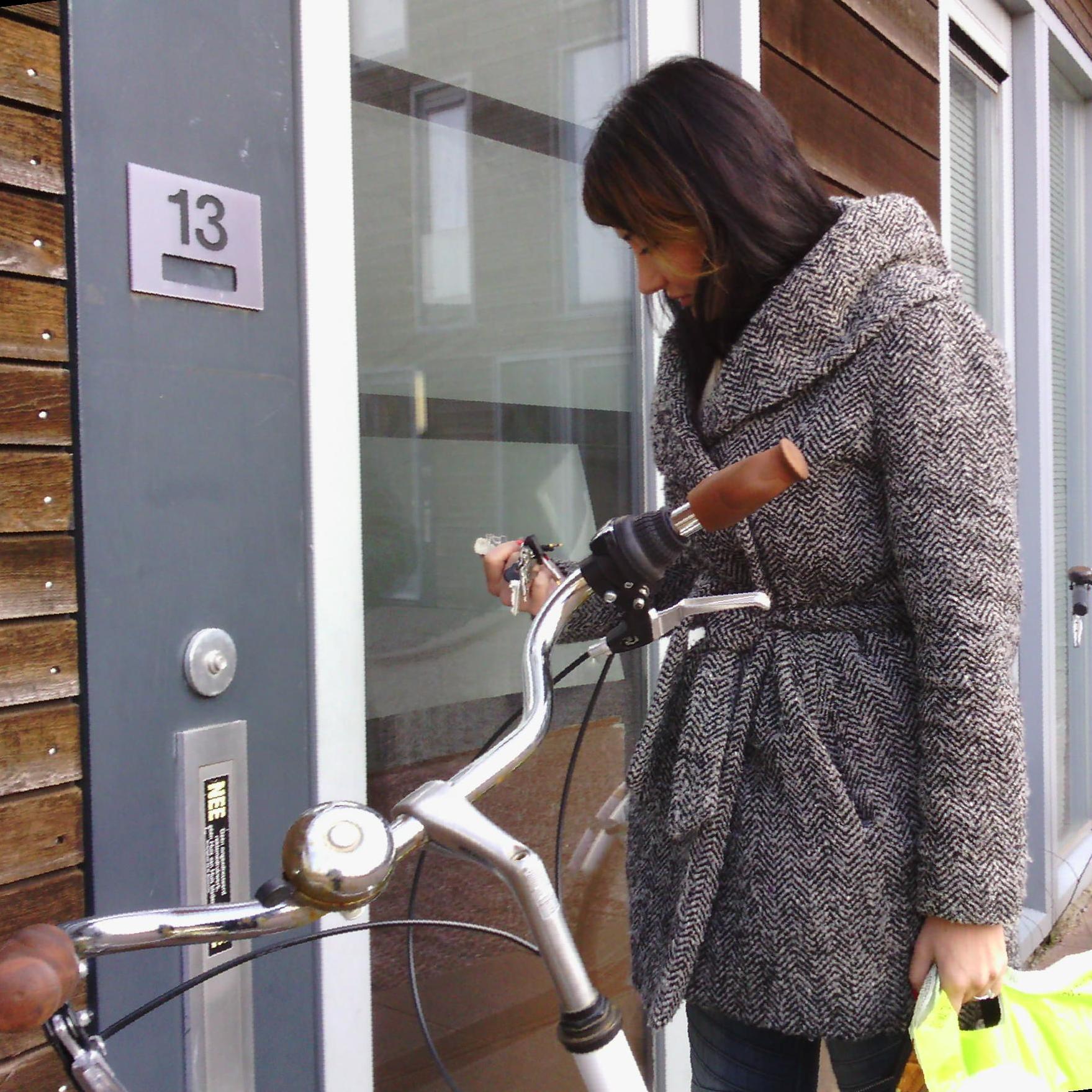} 
 \caption{Rich images}
 \label{fig:rich_images}
 \end{subfigure}\hfill%
 \begin{subfigure}{.5\textwidth}
 \includegraphics[width=.49\textwidth]{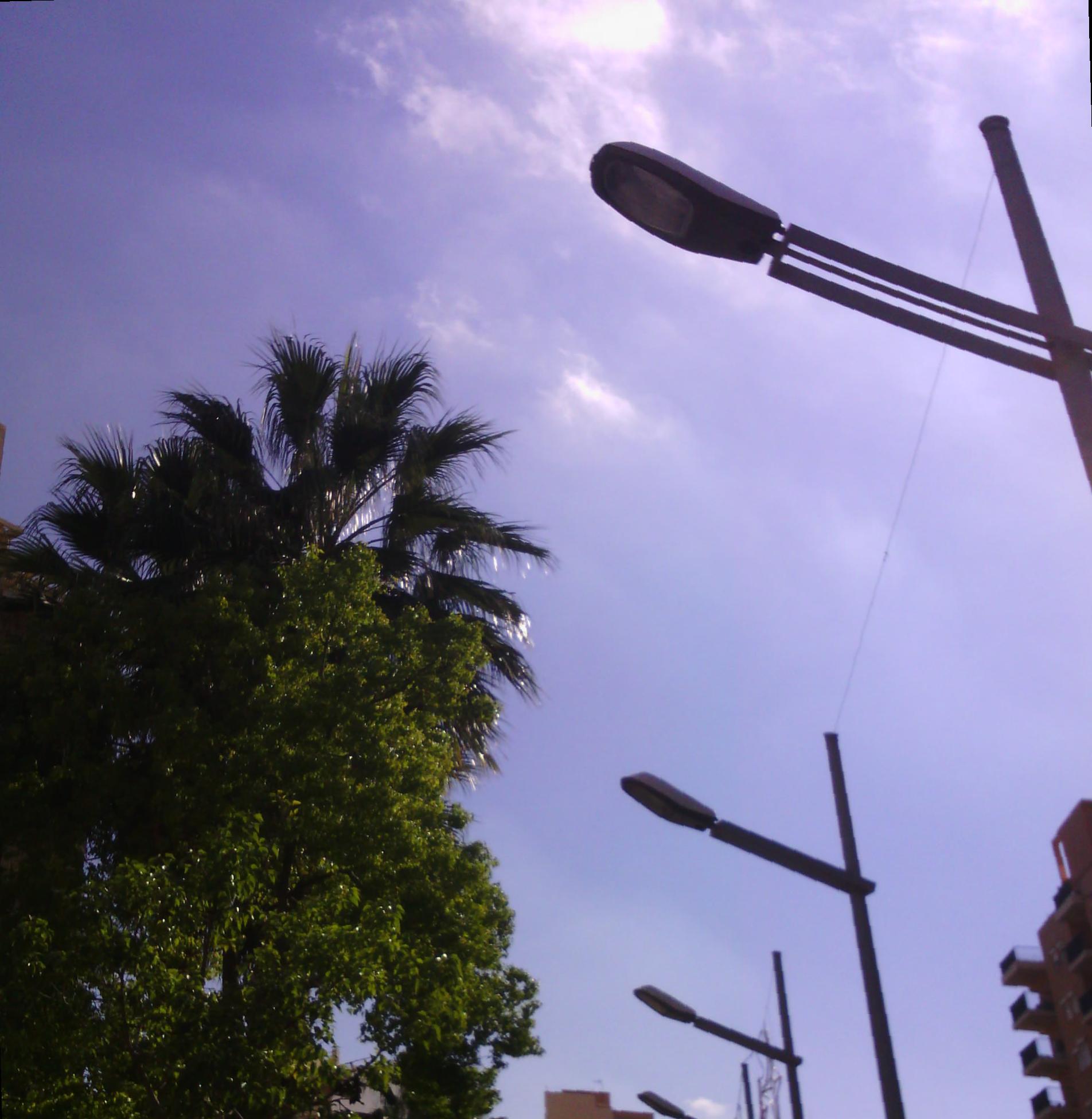}
 \includegraphics[width=.49\textwidth]{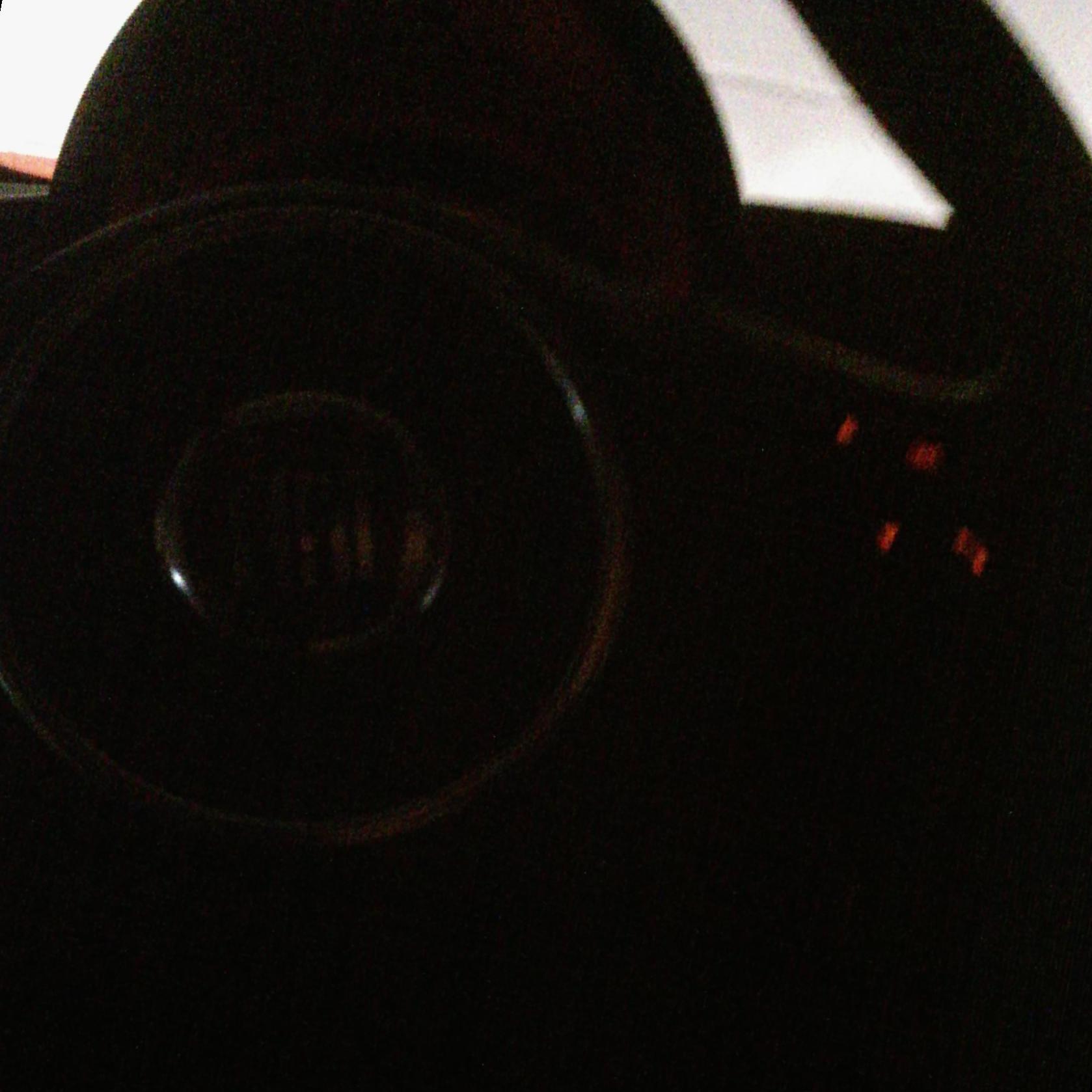} 
 \caption{Non-rich images}
 \label{fig:nonrich_images}
 \end{subfigure}
\end{figure}

In Fig. \ref{fig:rich_images} we can see some examples of egocentric rich images and in Fig. \ref{fig:nonrich_images} non-rich images. We observe that rich images show people or recognizable places. However, non-rich images are meaningless or dark images (that can hardly be seen), including pictures of the sky, ceilings or floor.

The resulting data was divided in training, validation, and test. Considering the pictures taken during the same day can be very similar, we proceeded to randomly separate the different days into the three different sets. First, the training set consists of 60\% of the days, in this case 9. Second, 20\% of the days, in this case 3, were defined as the validation set. Finally, the remaining 20\% was used for the test set.

\subsubsection{Evaluation Metrics:}
In order to evaluate the different results and compare them to get the best one, we make use of the F1-score (or F-measure) metric: $$F1 = 2*\frac{1}{\frac{1}{precision} + \frac{1}{recall}} = 2*\frac{precision * recall}{precision + recall}$$
where $precision$ is the quotient between the number of True Positives objects and the number of predicted positive elements; and $recall$ is the quotient between the number of True Positives objects and the number of real positive elements.\bigskip

\subsubsection{Quantitative Results:}

Currently, there are no previous works addressing the challenge we introduce in this work. Thus, in order to compare the performance of our proposed model, we have defined and compared several variations to our main pipeline (see results in Table \ref{table:resultscomparison}). 

As an alternative to our proposed approach (1), we tested an alternative feature vector representation by means of using the (2) Word2Vec word embedding \cite{goldberg2014word2vec}. This word characterization is a 300-dimensional vector representation created by Google that represents words in space depending on their semantic meaning (i.e. words with similar definitions will be represented close in space). The Word2Vec representation was used in two ways. On the one hand was used for defining the set of concepts related to "person" in the feature described as "Does the cell contain people?". Thus, we computed the similarity between the word "person" and any other concept detected in the image by the object detection and the maximum similarity achieved was used as an alternative to a 0/1 representation. On the other hand, the feature described as "Object Class" was replaced by the 300-dimensions Word2Vec representation.

In the test setting (3) we additionally applied a PCA dimensionality reduction to the Word2Vec representation. Finally, in (4) we used a Support Vector Machine (SVM) classifier instead of a Random Forest Classifier. We applied a Grid Search on the variables $C$ and $gamma$ for parameter selection over the validation set.

\begin{table}[h]
\centering
\begin{tabular}{llll}
 & \large\textbf{Precision} & \large\textbf{Recall} & \large\textbf{F1-score} \\ \hline
\textbf{(1) RFC} & 0.79 & 0.79 & 0.79 \\ 
\textbf{(2) RFC + Word2Vec} & 0.78 & 0.78 & 0.78 \\  
\textbf{(3) RFC + Word2Vec + PCA} & 0.74 & 0.75 & 0.75 \\ 
\textbf{(4) SVM} & 0.68 & 0.67 & 0.68\\
\end{tabular}
\caption{Comparison of the results}
\label{table:resultscomparison}
\vspace{-2em}
\end{table}

In conclusion we can see that using an RFC classifier (1) obtains better results than SVM (4) and at the same time none of the Word2Vec representations (2) and (3) helped improving the base results.

\subsubsection{Qualitative Results:}

\begin{figure}[ht!]
\centering    
\includegraphics[width=1\textwidth]{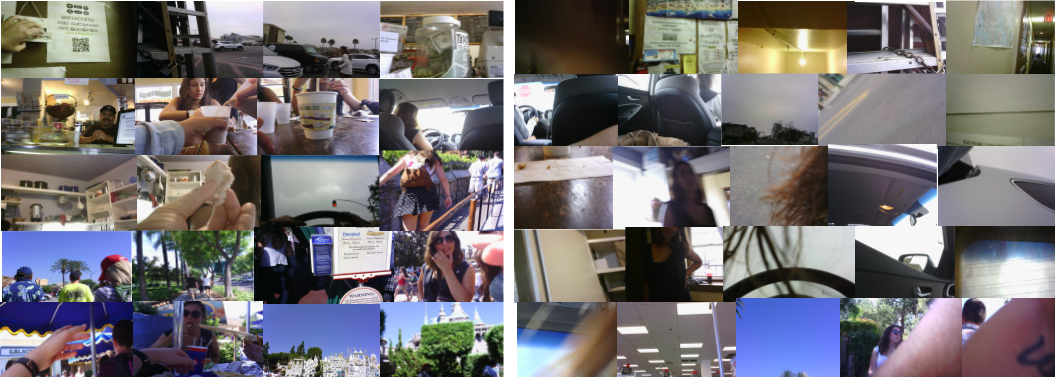}
    \caption{Example of rich (left) images selection, vs non-rich images rejection. From an egocentric photostream composed by 972 images, 221 were considered rich.}
\label{fig:examplesResults}
\end{figure}

Examples of the selected images by the proposed algorithm are shown in Fig. \ref{fig:examplesResults}. On one hand, we can observe that rich images (left) are clearer, without shadows and with people or focused objects, which allows the user to infer what is happening in the scene.  On the other hand, non-rich images (right) are discarded since they are not illustrative and make difficult the scene interpretation. 

Images selected by the proposed model are rich in information and memory trigger. We can foresee that the proposed model cannot only be used for serious games images selection, but also as a tool for images selection for autobiographical memories creation.

\section{Conclusions}
\label{section:conclusionsandfuture}
In this work, we have introduced a novel type of wearable computing application, aiming  to provide non pharmacological treatment for MCI patients and to improve their life quality. We discussed lifelogging pictures obtained from wearable cameras combined with serious games as a channel for personalized treatments. We also introduced and tested a novel computer vision technique to classify rich and non rich images obtained from first-person point of view. We obtain 79\% F1-score, promising results that will be further studied. 

As future work, we will implement more serious games to be included in the application tool. Specialists will use it for MCI patients, aiming to prove the the memory reinforcement hypothesis introduced in this work, as well as the motivation experienced by the subjects increase when using personalized rich images and serious games. Furthermore, in \cite{Sentiment2017Talavera}, positiveness from egocentric images was addressed. 
Moreover, we will go deeper on the analysis of users acceptance over the proposed technology, their willingness to use it, and the factors that determine their acceptance toward it. Further improvements of the methodology will be developed in order to obtain more accurate results. 

\vspace{-1em}
\section*{Acknowledgements}
This work was partially founded by Ministerio de Ciencia e Innovaci\'on of the Gobierno de Espa\~na, through the research project TIN2015-66951-C2. SGR 1219, CERCA, \textit{ICREA Academia 2014}, Grant 20141510 (Marat\'{o} TV3) and Grant FPU15/01347. The funders had no role in the study design, data collection, analysis, and preparation of the manuscript. The authors gratefully acknowledge the support of NVIDIA Corporation with the donation of the Titan Xp GPU used for this research.

\bibliographystyle{splncs03}
\bibliography{bibliographyfinal}

\end{document}